# TF-TransUNet1D: Time-Frequency Guided Transformer U-Net for Robust ECG Denoising in Digital Twin

Shijie Wang[1], Lei Li[1, *]

[1] Department of Biomedical Engineering, National University of Singapore, Singapore
`lei.li@nus.edu.sgu`

**Abstract.** Electrocardiogram (ECG) signals serve as a foundational data source for cardiac digital twins, yet their diagnostic utility is frequently compromised by noise and artifacts. To address this issue, we propose TF-TransUNet1D, a novel one-dimensional deep neural network that integrates a U-Net-based encoder–decoder architecture with a Transformer encoder, guided by a hybrid time–frequency domain loss. The model is designed to simultaneously capture local morphological features and long-range temporal dependencies, which are critical for preserving the diagnostic integrity of ECG signals. To enhance denoising robustness, we introduce a dual-domain loss function that jointly optimizes waveform reconstruction in the time domain and spectral fidelity in the frequency domain. In particular, the frequency-domain component effectively suppresses high-frequency noise while maintaining the spectral structure of the signal, enabling recovery of subtle but clinically significant waveform components. We evaluate TF-TransUNet1D using synthetically corrupted signals from the MIT-BIH Arrhythmia Database and the Noise Stress Test Database (NSTDB). Comparative experiments against state-of-the-art baselines demonstrate consistent superiority of our model in terms of SNR improvement and error metrics, achieving a mean absolute error of 0.1285 and Pearson correlation coefficient of 0.9540. By delivering high-precision denoising, this work bridges a critical gap in pre-processing pipelines for cardiac digital twins, enabling more reliable real-time monitoring and personalized modeling.

## 1 Introduction

Accurate electrocardiogram (ECG) signal denoising is a foundational prerequisite for cardiac digital twin systems, where high-fidelity physiological data drives personalized modeling and real-time diagnostics [1, 2]. While conventional signal processing techniques, such as linear filtering, Gaussian smoothing, and wavelet transforms [3], have been widely adopted for signal enhancement [4, 5], their rigidity often distorts morphological features critical to digital twin accuracy (e.g., P-wave duration or ST-segment slope). Advanced methods like adaptive filtering and Empirical Mode Decomposition (EMD) [6–8] improve adaptability to non-stationary noise but fail to preserve spectral integrity when noise overlaps with the diagnostic bandwidth of ECG. Such limitations introduce cascading errors in digital twin pipelines, where even minor waveform alterations can compromise predictive simulations of arrhythmias or drug responses.

In recent years, the application of deep learning methods to ECG signal denoising has achieved significant progress. For example, Autoencoders have been used for ECG denoising by reconstructing clean signals from noisy inputs [9, 16]. Deep Recurrent Neural Networks (RNNs) have been utilized for this purpose [10], and multi-branch Convolutional Neural Networks (CNNs), such as DeepFilter, have demonstrated excellent performance in removing specific types of noise. Despite these advancements, a fundamental challenge persists how to preserve local signal features while simultaneously capturing the global context. This difficulty arises because conventional CNNs or autoencoders [11, 16] typically operate with a limited receptive field, which restricts their ability to capture long-range dependencies. Conversely, Transformer-based models, [13, 15] excel at global modeling but may lose critical local details if they lack multi-scale feature pathways. Furthermore, neural networks [14, 17] often overlook the preservation and reconstruction of spectral information [18], which can lead to distortion in the recovered ECG signal. The classic TransUNet architecture incurs high computational cost at each CNN encoder and decoder stage.

Motivated by these challenges, we have designed a novel and lightweight TF-TransUNet1D, a lightweight yet powerful architecture specifically designed to meet the stringent signal fidelity requirements of cardiac digital twin systems. Our novel hybrid architecture combines the complementary strengths of local feature representation and global context aggregation through two key innovations: (1) a time-frequency constrained Transformer U-Net that simultaneously preserves both morphological details and spectral characteristics, and (2) a multi-domain loss function that jointly optimizes temporal waveform accuracy and spectral fidelity. This dual-domain optimization ensures clinically-relevant features (e.g., ST-segment morphology and P-wave characteristics) remain intact during denoising, which is crucial for reliable cardiac digital twins.

## 2   Methodology

### 2.1   Temporal-Aware Encoder-Decoder with Skip Connections

The encoder progressively compresses the input ECG feature representations. At the deepest level, after the D4 convolutional block, the feature map is fed into the Transformer module. Positional encoding is added to preserve temporal order, and the sequence is processed by $N$ stacked Transformer encoder layers, each consisting of multi-head self-attention and feedforward sub-layers. The enhanced feature sequence is then reshaped (if necessary) back into spatial layout and passed to the U-Net decoder.

The decoder consists of Upsampling layers (transposed convolutions) U1–U4, each followed by a double convolution block. Skip connections from corresponding encoder stages are concatenated with the Upsampled feature maps (e.g., U1 output is concatenated with features from D3), helping to retain high-frequency ECG details that may have been lost during compression. The final layer is a 1D convolution with kernel size 1 (OutConv), which maps the decoded high-dimensional features back to the target output channels—1 for single-lead ECG denoising. TF-TransUNet1D takes input segments of shape (batch size, 1, 3600) and outputs denoised ECG segments of the same shape. All convolution layers use appropriate padding to maintain alignment and apply batch normalization to aid convergence (unless otherwise specified).

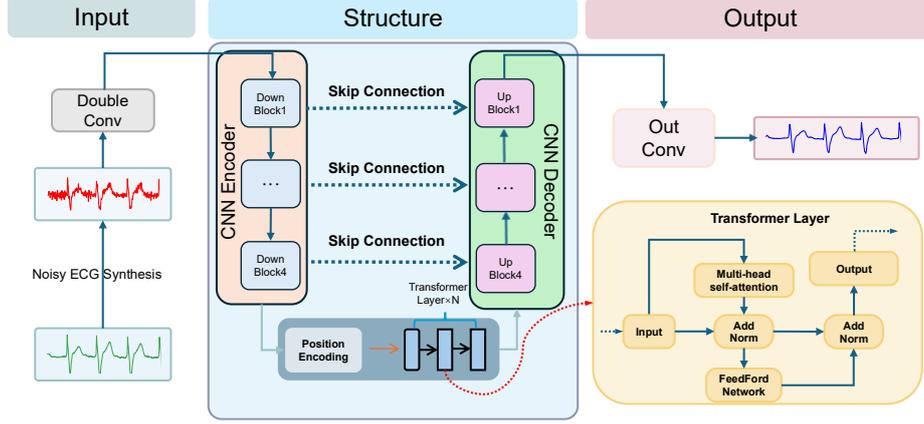

**Fig. 1.** A TF-TransUNet1D architecture, which follows a U-Net-style encoder-decoder design with four DownSampling/Upsampling levels and a Transformer encoder at the bottleneck. The encoder includes an input convolution block (INC) and three down sampling stages (D1–D4), each with two 1D convolutional layers (ReLU) and a max-pooling layer.

### 2.2 Modeling Long-Range Dependencies via Transformer

The Transformer encoder integrated into TF-TransUNet1D is designed to model long-range dependencies in ECG sequences. Each encoder layer follows the standard architecture: Multi-Head Self-Attention (MHSA) followed by a position-wise feedforward network. Formally, given an input sequence of feature vectors $X \in \mathbb{R}^{T \times d}$, where T is the sequence length and d is the feature dimension per token, the self-attention mechanism first computes query, key, and value matrices via learned linear projections of X. The self-attention output for a single head is computed as shown in Equation (1), where a context-aware representation is produced by taking the weighted average of value vectors, with weights determined by the similarity between queries and keys. In the multi-head setting, attention is computed in parallel across H heads, and the results are concatenated and linearly transformed back to d-dimensions.

Residual skip connections and layer normalization (LN) are applied around both the MHSA and feedforward sub-layers. At the heart of the TransUNet-1D architecture, a standard Transformer encoder is embedded within the bottleneck of the U-Net. This design strategically combines the strong local feature extraction of CNN with the Transformer's ability to model global context. The U-Net encoder converts the raw ECG signal (length L) into a compact feature sequence of length $L' \ll L$ and channel size $c' \gg 1$. While rich in semantic information, this sequence's receptive field remains limited by the depth of CNN layers.

To overcome this, the Transformer encoder takes the compressed sequence as input and applies Multi-Head Self-Attention (MHSA) to model long-range dependencies. Each time step attends to all others, enabling the model to learn global temporal relationships—crucial for understanding the dynamics between QRS complexes, P waves, and T waves across the full 10-second segment. This global context helps distinguish physiological signals from non-stationary noise such as muscle artifacts. Through iterative refinement, the Transformer enhances meaningful patterns while suppressing irrelevant noise, yielding a purified feature representation for the decoder to reconstruct high-quality ECG signals.

### 2.3   Dual-Domain Loss Guided Strategy

To effectively guide the training of the TransUNet-1D model, we designed and adopted a multi-objective loss function that jointly optimizes both time-domain waveform reconstruction accuracy and frequency-domain spectral consistency. The total loss $L_{loss}$ is a weighted sum of these two components. Let $y[n]$ denote the clean ECG signal and $\hat{y}[n]$ the denoised output from the model (for n=1, …, $N$ samples within a segment).

To balance convergence stability and robustness to artifacts, we ultimately chose the Smooth L1 loss. This loss behaves like mean squared error (MSE) when the absolute error $|y - \hat{y}| < \beta$, providing smooth gradients for precise convergence, and like mean absolute error (MAE) when $|y - \hat{y}| \geq \beta$, applying a linear penalty to reduce the influence of outliers and prevent overreactions to noise. This design allows the model to accurately reconstruct the underlying ECG morphology while remaining resilient to residual noise and prediction errors, resulting in more natural and reliable denoised signals.

In addition to time-domain fidelity, we introduce a spectral loss $L_{\text{spectral}}$ to ensure that the denoised signal retains the correct frequency content compared to the ground truth. We compute the Fast Fourier Transform (FFT) of both signals and compare their magnitudes:

$$\mathcal{L}_{spectral} = \frac{1}{K}\sum_{k=1}^{K}\left(|Y[k]| - |\hat{Y}[k]|\right)^2, \tag{4}$$

Where $Y[k]$ and $\hat{Y}[k]$ are the discrete Fourier magnitudes of the clean and reconstructed signals at frequency bin $k$, and $K$ is the total number of frequency bins.

This term penalizes the discrepancy between the power spectra of $\hat{y}$ and $y$, guiding the model to not only minimize point-wise errors but also correctly redistribute energy across frequencies. This is crucial for preserving characteristic ECG bandwidths and preventing distortion of features such as QRS frequency components. The total training loss is a weighted sum of the two components:

$$\mathcal{L}_{total} = w_{time}\mathcal{L}_{time} + w_{spectral}\mathcal{L}_{spectral} \tag{5}$$

By optimizing $L_{\text{total}}$, TF-TransUNet1D learns to produce outputs that closely match the clean ECG in both the time and frequency domains. Incorporating $L_{\text{spectral}}$ is especially beneficial under heavy noise conditions, as it prevents the network from simply smoothing the signal, a strategy that may minimize MSE but also attenuate important high-frequency components. Instead, the model is encouraged to recover the true frequency characteristics of the signal, resulting in denoising that is more physiologically accurate.

## 3  Experiments

### 3.1  Experiments Materials

**Noisy Signal Synthesis and Preprocessing.** Clean ECG signals from the MIT-BIH Arrhythmia Database are mixed with various noise types from the Noise Stress Test Database (NSTDB) to generate paired clean/noisy ECG segments with specific signal-to-noise ratios (SNRs). A sliding window mechanism is employed to divide the recordings into multiple shorter, fixed-length segments (each input segment comprises 3,600 sampling points, corresponding to a duration of 10 seconds) for model training and evaluation. Noise sources include baseline wander, muscle (EMG) noise, electrode motion artifacts, and powerline interference. All segments are z-normalized (zero mean, unit variance), and the same random noise instance is added to the corresponding clean segment to form paired training data. A portion of the dataset is reserved for validation to tune hyperparameters (e.g., loss weights, learning rate) and implement early stopping. The test set consists of ECG segments from subjects not seen during training, ensuring a fair assessment of the model's generalization ability.

**Implementation.** The deep learning framework was implemented from scratch using the PyTorch library, with custom modules defining the model architecture (model.py), data processing pipelines, and training procedures. The training process utilized AdamW with cosine annealing decay ($T\_max = 100, \eta\_min = 1e − 6$) for optimization. All models were trained for 100 epochs with a batch size of 16 on NVIDIA GeForce RTX 4090GPU.

**Evaluation Metrics.** The performance of 1D ECG signal denoising is evaluated using standard metrics that assess both reconstruction fidelity and noise suppression. Signal-to-Noise Ratio (SNR) measures how effectively noise is reduced—higher values indicate better preservation of the original ECG waveform. SNR Improvement (SNRI), the difference between output and input SNR, reflects the model's noise reduction capability. Percentage Root Mean Square Difference (PRD) quantifies the normalized reconstruction error, with lower values indicating better retention of the signal's power. Pearson Correlation Coefficient (PCC) evaluates morphological similarity, showing how well the ECG waveform shape is preserved. Mean Absolute Error (MAE) measures the average amplitude difference between clean and denoised signals, reflecting point-wise reconstruction accuracy. Together, SNRI reflects noise suppression, PCC captures waveform integrity, and MAE assesses numerical precision—providing a comprehensive evaluation of ECG denoising performance.

### 3.2 Results

**Quantitative Results Analysis.** To evaluate the performance of the model, we conducted experiments using our synthesized noisy dataset on the proposed model as well as several commonly used baseline models, under the same settings such as initial learning rate, optimizer, and other parameters.

Table 1 presents a comparison of denoising performance using different models under the condition of mixed noise with a target SNR of 0 dB. Our proposed TF-TransUNet1D model, constrained by a time-frequency hybrid loss function, achieves the best results across all evaluation metrics, including MAE, PCC, and SNRI. In particular, in terms of Mean Absolute Error (MAE), our model reduces the error by approximately 24.3% compared to the second-best performing model, indicating that the average amplitude difference between the reconstructed and clean signals is minimal. Furthermore, combined with the high Pearson Correlation Coefficient, the results highlight the strong performance of the TF-TransUNet1D model guided by Dual-Domain Loss in preserving signal fidelity and spectral characteristics.

**Table 1.** Model Comparison for Denoising Tasks with Three-Type Signal Mixtures at 0 dB

| Model | MAE | PCC | SNRI |
|---|---|---|---|
| CNN-LSTM | 0.1819 | 0.916 | 10.61 |
| U-Net 1D | 0.1697 | 0.9195 | 10.85 |
| FastRNN | 0.2057 | 0.9138 | 9.74 |
| TF-TransUNet1D | 0.1285 | 0.9540 | 13.36 |

Figure 2 presents systematic experimental results on a synthesized noisy dataset, demonstrating the model's strong robustness and generalization when handling diverse single and combined noise types. Under the extreme low Signal-to-Noise Ratio (SNR = 0 dB), the model achieved the lowest Mean Absolute Error (MAE) of 0.0819 against Motion Artifact (MA) noise and the highest Pearson Correlation Coefficient (PCC) of 0.981 against Burst Noise (BW). When the SNR increased to 10 dB, the PCC further improved to 0.992, indicating high waveform reconstruction quality under mixed-frequency interference. While slight performance degradation occurred with composite noise (BW + Electromagnetic Interference (EM) + MA), the model's MAE and PCC remained within acceptable ranges. It consistently outperformed in Signal-to-Noise Ratio Improvement (SNRI), affirming the effectiveness of the dual-branch architecture and hybrid loss design in complex noise scenarios.

**Qualitative Analysis of Results.** The qualitative visualization of ECG denoising using

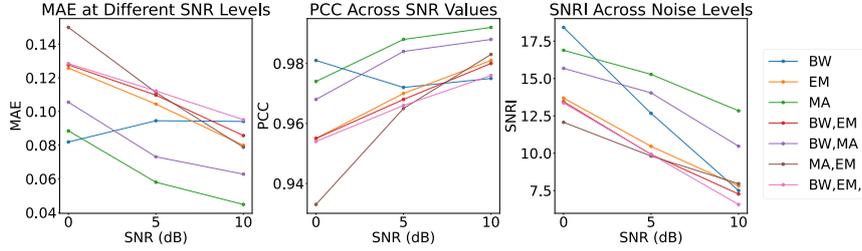

**Fig. 2.** Three-line plots depicting model performance across different noise types and target SNR levels. The figure consists of three subplots, each showing one evaluation metric (MAE, PCC, and SNRI) as a function of SNR (0 dB, 5 dB, and 10 dB). Each line corresponds to a specific noise type or noise combination, with a shared legend positioned on the right.

TF-TransUNet1D (Fig. 3) demonstrates marked improvements in waveform morphology preservation and noise suppression. The model effectively smooths high-frequency myoelectric artifacts and attenuates substantial motion artifacts, restoring normal cardiac rhythm while maintaining diagnostically critical features—distinct P-waves, sharply defined QRS complexes, and well-preserved T-waves—with exceptional morphological, amplitude, and phase fidelity to gold-standard references. This success is largely attributed to the incorporation of a frequency-domain loss component, which overcomes the limitations of traditional time-domain methods by ensuring spectral consistency, particularly within powerline interference bands, thereby preserving the integrity of clinically important waveforms. Comprehensive quantitative and qualitative evaluations confirm that TF-TransUNet1D delivers superior denoising performance and robust generalization for ECG signals. Furthermore, its lightweight design facilitates real-time deployment in clinical applications such as ambulatory monitoring and wearable medical devices.

## 4    Conclusion

In this work, we present TF-TransUNet1D, a Transformer-enhanced U-Net model designed for denoising 1D ECG signals. By integrating global self-attention with local feature extraction and skip connections, the architecture effectively bridges convolution- and attention-based approaches. It achieves robust noise reduction across various artifact types and noise levels, while preserving the morphological integrity of ECG waveforms. Driven by a multi-objective loss function that balances time-domain precision with frequency-domain fidelity, TF-TransUNet1D delivers denoised outputs that are both quantitatively accurate and qualitatively faithful. Importantly, this work offers an efficient and scalable strategy for incorporating Transformer blocks into biomedical time-series models, with minimal architectural overhead. By enhancing ECG signal quality, TF-TransUNet1D supports more reliable feature extraction and physiological modeling, making it a valuable component for building high-fidelity cardiac digital twins. The proposed framework contributes to the broader vision of personalized cardiovascular care, where accurate, artifact-free ECG data forms the foundation for dynamic patient-specific digital twin simulations.

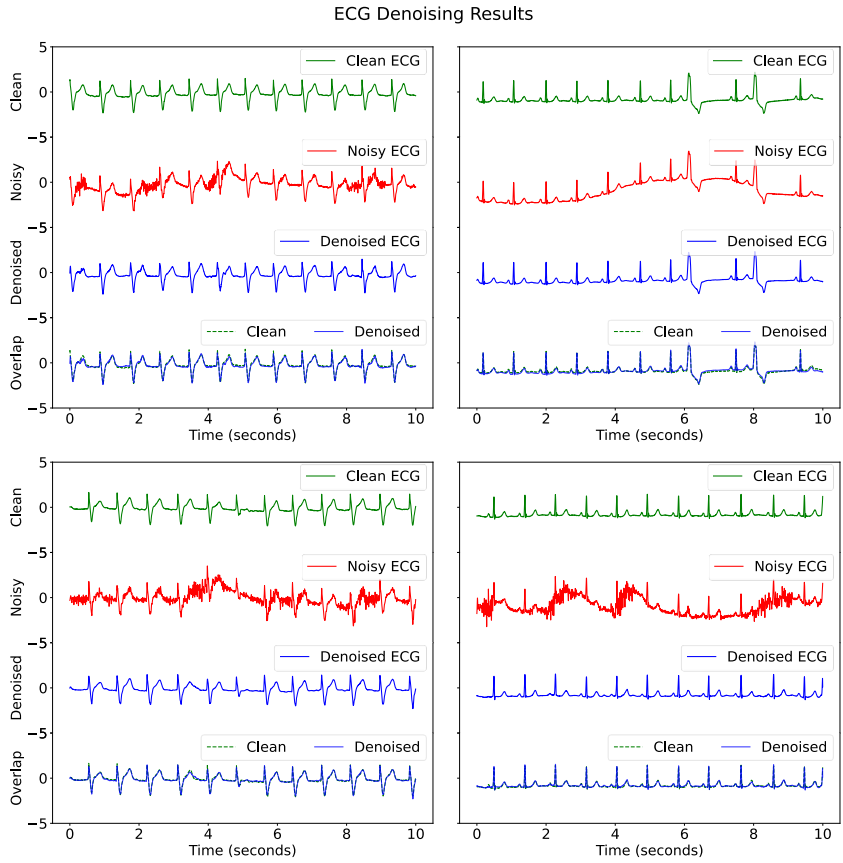

**Fig. 3.** Denoising Results of ECG Signals Using the TF-TransUNet1D Model

## Acknowledgment

This work was supported by NUS start-up funding to L. Li. We also thank the PhysioNet team for providing access to the MIT-BIH Arrhythmia Database.